%% file: main.tex
\documentclass[runningheads]{llncs}

\usepackage{eccv}

\usepackage{eccvabbrv}

\usepackage{graphicx}
\usepackage{booktabs}

\usepackage[accsupp]{axessibility}  % Improves PDF readability for those with disabilities.

\usepackage{hyperref}

\usepackage{orcidlink}

\usepackage{makecell}
\usepackage{booktabs,multirow,adjustbox,diagbox,threeparttable}

\input{math_commands}

\begin{document}

\title{Paying More Attention to Image: A Training-Free Method for Alleviating Hallucination in LVLMs} 

\titlerunning{PAI}

\author{
  Shi Liu\inst{1} \and
  Kecheng Zheng\inst{1,2}$^\dag$ \and
  Wei Chen\inst{1}$^\dag$
}

\authorrunning{Liu et al.}

\institute{State Key Lab of CAD\&CG, Zhejiang University \and
Ant Group \\
\email{\{liushi0927,zkechengzk\}@gmail.com} \\
\email{chenvis@zju.edu.cn}}

\maketitle
\footnotetext[2]{$^\dag$ Corresponding authors.}

\input{sections/0.abs.tex}
\input{sections/1.intro.tex}
\input{sections/2.related.tex}
\input{sections/3.method.tex}

\input{sections/4.exp.tex}
\input{sections/5.conclusion.tex}
\section*{Acknowledgements}
This work is supported by the National Science Foundation of China (62132017), Zhejiang Provincial Natural Science Foundation of China (LD24F020011).
\input{sections/6.ref.tex}
\input{sections/7.appendix.tex}

% \bibliography{main}
\end{document}

%% file: math_commands.tex
\usepackage{amsmath,amsfonts,bm}

\def\eqref#1{equation~\ref{#1}}
\def\1{\bm{1}}

\def\vx{{\bm{x}}}
\def\vy{{\bm{y}}}

\def\mA{{\bm{A}}}

\def\mK{{\bm{K}}}

\def\mO{{\bm{O}}}

\def\mQ{{\bm{Q}}}

\def\mV{{\bm{V}}}

\def\mX{{\bm{X}}}

\DeclareMathAlphabet{\mathsfit}{\encodingdefault}{\sfdefault}{m}{sl}
\SetMathAlphabet{\mathsfit}{bold}{\encodingdefault}{\sfdefault}{bx}{n}

\newcommand{\pmodel}{p_{\rm{model}}}

\newcommand{\R}{\mathbb{R}}

\newcommand{\softmax}{\mathrm{softmax}}

%% file: sections/0.abs.tex
\begin{abstract}
Existing Large Vision-Language Models (LVLMs) primarily align image features of vision encoder with Large Language Models (LLMs) to leverage their superior text generation capabilities.
However, the scale disparity between vision encoder and language model may led to LLMs assuming a predominant role in multi-modal comprehension.
This imbalance in LVLMs may result in the instances of hallucinatory. 
Concretely, LVLMs may generate consistent descriptions with or without visual input, indicating that certain outputs are influenced solely by context text.
We refer to this phenomenon as ``text inertia.''
To counteract this issue, we introduce a training-free algorithm to find an equilibrium point between image comprehension and language inference. 
Specifically, we adaptively involve adjusting and amplifying the attention weights assigned to image tokens, thereby granting greater prominence to visual elements.
Meanwhile, we subtract the logits of multi-modal inputs from ones of pure text input, which can help LVLMs be not biased towards LLMs.
By enhancing images tokens and reducing the stubborn output of LLM, we can let LVLM pay more attention to images, towards alleviating text inertia and reducing the hallucination in LVLMs.
Our extensive experiments shows that this method substantially reduces the frequency of hallucinatory outputs in various LVLMs in terms of different metrics.
Project page is available at~\url{https://lalbj.github.io/projects/PAI/}
\end{abstract}

%% file: sections/1.intro.tex
\section{Introduction}\label{sec:intro}

Recently, Large Vision-Language Models (LVLMs) have made significant strides, exhibiting impressive capabilities across a multitude of tasks~\cite{lenna,chartllama,lisa,prompthighlighter}. 
However, these models still struggle with the phenomenon of hallucination. 
Specifically, there is often a mismatch between the textual content generated by the model and the actual visual input it receives~\cite{lvlms_survey}.

\begin{figure}[t!]    
    \centering    
    \includegraphics[width=\linewidth]{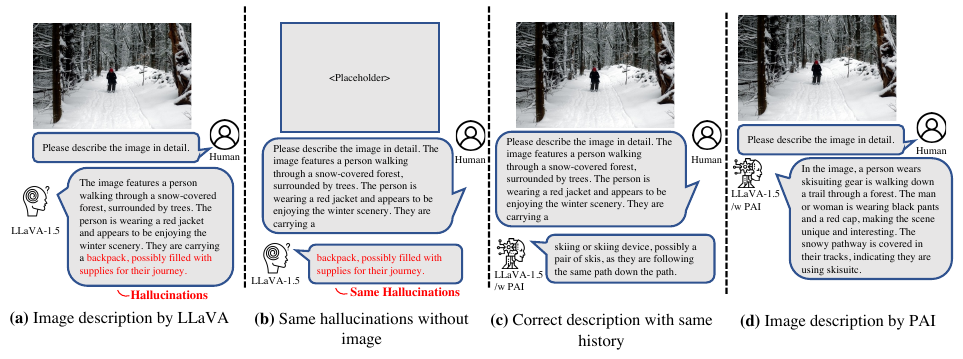}    
    \caption{We present an examination of various input settings, with hallucinations specifically highlighted in \textcolor{red}{red}. (a) When using LLAVA for image description, it generates a hallucinated description. (b) Even without image input, when only the historical response preceding the hallucinated description is input to LLAVA, it reproduces the same hallucinated description, a phenomenon we refer to as ``Text Inertia''. (c) Our proposed method, PAI, effectively mitigates this text inertia problem and yields accurate descriptions. (d) Utilizing PAI for image description results in a significantly more precise description.}    
      
    \label{fig:teaser}    
\end{figure}

\begin{figure}[t!]  
  \begin{minipage}[b]{0.49\linewidth}  
    \centering  
    \includegraphics[width=\linewidth]{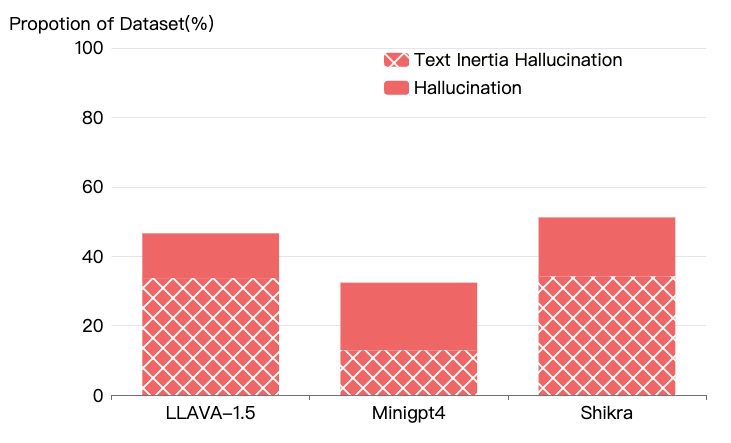}  
    \caption{Percentage of text inertia hallucination in all hallucination (calculated with 500 samples). For specific calculation processes, please refer to the supplementary material section A.}  
    
    \label{fig:barchart_stat}  
  \end{minipage}  
  \hfill  
  \begin{minipage}[b]{0.49\linewidth}  
    \centering  
    \includegraphics[width=\linewidth]{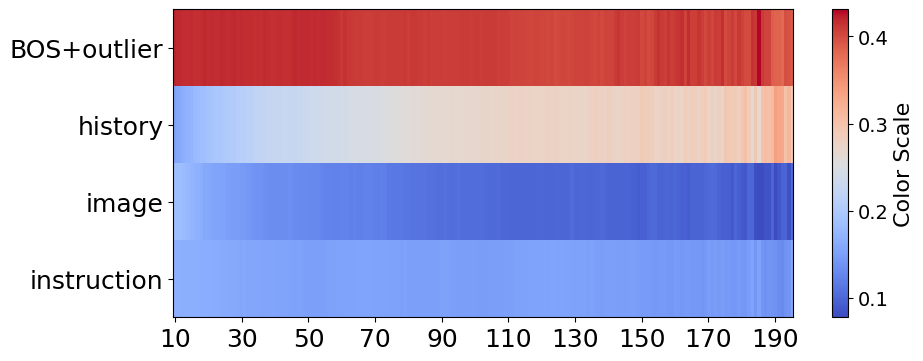}  
    \caption{Visualization of the average attention ratio of different content. The x-axis denotes the sequence length of the history tokens. The lengths of the image, instruction, BOS, and outlier tokens~\cite{outlier} are all fixed as they are part of the model input, with 576 tokens for the image, 21 for the instruction, and 1 each for the BOS and outlier tokens.}  
    
    \label{attn_ratio_content}  
  \end{minipage}  
\end{figure} 

Hallucination in LVLMs is often attributed to issues with modality alignment, leading to the development of mitigating strategies through alignment training optimization~\cite{align1,archit,archit2}. 
However, is hallucination in LVLMs merely a result of the model's capacity, and can it only be alleviated through additional training?
We propose a scenario where LVLMs generate a hallucinated object description. 
Specifically, even when the image input is removed and only the generated text preceding the hallucinated object word is retained, the LVLMs persist in producing the same hallucinated description, as depicted in ~\cref{fig:teaser}.     

In order to empirically investigate this behavior, we conducted tests on three LVLMs within the context of image describing tasks on the COCO dataset. 
We identified and conducted a statistical analysis on instances where LVLMs generated identical hallucinated object descriptions, even when the input was exclusively historical response text without any image.
The observation from \cref{fig:barchart_stat} clearly indicates that, even with the application of rigorous identification settings, the phenomenon continues to represent a substantial proportion.

We referred this phenomenon as \textbf{``Text Inertia''}. 
Our hypothesis is that text inertia arises due to the current generative paradigms mapping image representations onto the text representation space as text tokens. 
In this mechanism, the LLM becomes the dominant character, and the inference process lacks additional handling of image tokens, leading to their neglect during the generation process. 
To validate this hypothesis, we have analyzed the attention values ratios of the LLaVA model during the inference process in \cref{attn_ratio_content}. 
Our findings show that despite image tokens occupying a significant proportion, they do not receive substantial attention under the current mechanism. 
This multimodal chat resembles more of an automatic completion based on context rather than a continuous attention to the image for completion.

To close this gap, we introduce a method refered to as \textbf{Pay Attention to Image (PAI)}. At a high level, PAI intervenes in the inference process to make it more image-centric, following the original image perception direction.
To achieve this, we focus on the self-attention heads in the decoder layers of LVLMs. 
We enhance the attention weights for image tokens in their original directions during inference. 
This allows us to use the updated attention matrix to calculate the hidden states for the generated token, thereby incorporating more consideration for image representation during the generation process. 
To further mitigate text inertia, we construct the input using instruction tokens and historical response tokens, and subtract the model logits of this input from the logits of original model with image tokens input. 
This strategy helps to reduce the influence of language priors during the generation process. 
Unlike previous methods for mitigating hallucination that require additional training or external tools, our approach is training-free. 
Moreover, we are the first to propose an inference intervention method for mitigating hallucination in LVLMs.

We then evaluate the response accuracy in the image description task from long sequence generation perspective by employing the CHAIR metric~\cite{chair} and GPT-4V. 
In addition, we use POPE~\cite{pope} and MMHal-Bench~\cite{mmhalbench} to more comprehensively evaluate the model's hallucination performance on VQA task. 
Furthermore, we have constructed single-turn and multi-turn chat evaluations for POPE.
Since our model intervenes in the inference process, it can be used for any decoding method. 
Therefore, we conducted experiments on three decoding methods of the three models. 
The experimental results proved the effectiveness of our method in mitigating hallucinations.

% In summary, the contributions can be summarized as follows:

% 1. We demonstrate the phenomenon of text inertia that LVLMs answer only by language context without looking at the image.

% 2. Based on the above analysis, we further found that the hallucination behavior is related to the neglect of images. Therefore, we further proposed a inference intervention algorithm to make LVLMs pay more attention to the image.

% 3. Through extensive experiments, we have demonstrated the effectiveness of PAI in mitigating hallucination problems in a training-free manner.

\begin{figure*}[t]  
    \centering  
    \includegraphics[width=\linewidth]{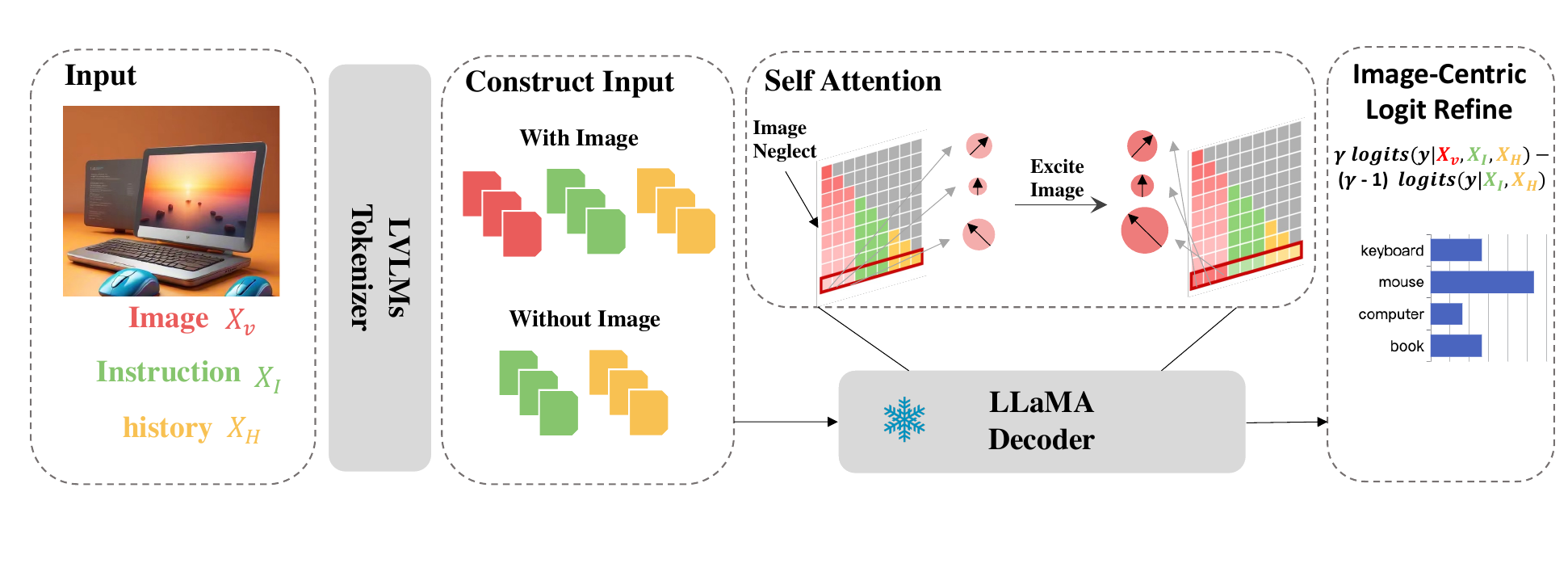}  
    \caption{
    Architecture of our PAI. 
    %
    % It demonstrates the inference process of a single token within our approach. 
    %
    To alleviate text inertia, we additionally construct an input without image. 
    Throughout the forward inference process, we amplify the focus on the image token by edit self-attention maps in LLaMA. 
    Ultimately, we subtract the logits distribution of the language prior during decoding to achieve an accurate description.
    }
    \label{fig:framework}  
\end{figure*} 

%% file: sections/2.related.tex
\section{Related Work}\label{sec:related}
\subsection{Large Vision-Language Models}
The development of pre-training techniques~\cite{instructgpt,gpt3} and instruction tuning techniques~\cite{instructtuning,instructtuning2} has rapidly advanced LLMs technology, such as LLaMA~\cite{llama} and Vicuna~\cite{vicuna}, further leading to the prosperity of LVLMs technology. Early works, like Flamingo~\cite{flamingo} and BLIP-2~\cite{blip2}, have successfully adapted LLMs to visual tasks, demonstrating notable generative capabilities and in-context learning abilities. Recently, the capabilities of LVLMs have further advanced under the influence of visual instruction tuning techniques~\cite{llava,llava-1.5}. Using different projectors to map images to the text domain, thereby endowing language generation models with image understanding capabilities, is also a hot research topic~\cite{llava,minigpt4,instructblip,mplugowl}. Additionally, several studies focus on visual language tasks such as grounding capabilities~\cite{shikra} and reasoning capabilities~\cite{lisa}. However, recent LVLMs still face the issue of hallucination generation~\cite{mitigating}.

\subsection{Mitigation of LVLMs Hallucination}
Hallucination in LVLMs refers to contradictions between the image input and the textual output. Various methods have been proposed to mitigate hallucination. The most direct reason for hallucination generation is that hallucination arises from data bias and knowledge gaps between vision and language. Therefore, better data filtering methods~\cite{ciem,lrv-instruct,ferret} and higher quality annotated data~\cite{dataset1} are introduced. Simultaneously, these methods also imply the need for more alignment training~\cite{align1} or adjustments in the model architecture~\cite{archit,archit2}. These methods can achieve good results, but they are time-consuming and require high computational resources.

Apart from addressing the ability of LVLMs itself, hallucination can also be mitigated through post-processing methods. This method usually involves using an additional module or external tools to edit the response. Recent methods such as LURE~\cite{lure} utilize additional data to train a state detector and when hallucination issues are detected, content is regenerated by a revisor model. Woodpecker~\cite{woodpecker} introduces an external visual model to inspect entities extracted from the response, and then the detection results are handed over to the generation model to regenerate better answers. These methods also extend the inference chain and increase inference costs.

Training-free hallucination mitigation methods have so far only been attempted in decoding methods. OPERA~\cite{opera} discovered an abnormal attention pattern that accompanies model decoding. It was statistically found that this pattern often accompanies hallucination descriptions, and thus a detection and mitigation method was proposed based on this pattern to alleviate the hallucination faced by the model. VCD~\cite{vcd} introduced the notion that visual uncertainty increases hallucination descriptions and, based on this discovery, proposed a contrast decoding method to alleviate hallucination issues.

%% file: sections/3.method.tex
\section{Preliminaries}

The architecture of LVLMs typically comprises of three main components: an image encoder, a projector, and a language decoder. Both the image encoder and language decoder are usually pre-trained. The image encoder is employed to transform images into image tokens, which are subsequently mapped to the text representation space by the projector. This process enables the concatenation of image tokens with text tokens that are then fed into the language decoder. The language decoder subsequently generates corresponding responses based on the provided instructions.

\textbf{The existing projectors.} 
Currently, projectors predominantly fall into two categories: linear projectors and resamplers. A projector takes N visual features from the image encoder and transforms them into M visual tokens. The linear projector employs a multilayer perceptron to transform all visual features, maintaining a one-to-one transformation which meaning that M equals N. In contrast, the resampler does not preserve all visual features but instead samples visual cues (M, where M $\textless$ N). For instance, Q-former~\cite{blip2} utilizes M learnable queries and Bert~\cite{bert} to extract information from visual features. Given that the knowledge of images during the generation process solely originates from the output image tokens of the projectors, our attention is concentrated on the image tokens post-projection, irrespective of their preceding modeling process.

\textbf{Autoregressive language decoders.} 
Nearly all LVLMs adopt LLaMA-family models as their language decoders, which employ the self-attention mechanism.
The visual tokens processed by the projector are concatenated with text tokens and fed into the LLaMA, which carries out the forward decoding process.
From the perspective of a single attention head in a single layer, each head repeatedly performs the following attention operation with the same input shape:
\begin{equation}
\mO_h = \mA_h \mV_h, \quad \mA_h = \softmax \left(\frac{\mQ_h \mK_h^{\top}}{\sqrt{d_k}}\right).
\end{equation}

Each attention head $h$ performs an attention operation using its own set of queries $\mQ_h \in \R^{n \times d_k}$, keys $\mK_h \in \R^{n \times d_k}$, and values $\mV_h \in \R^{n \times d_k}$, where $n$ represents the sequence length and $d_k$ represents the hidden dimensions.
The output $\mO_h \in \R^{n \times d_k}$ is modeled by multiplying $\mV_h$ and the attention weights $\mA_h \in \R^{n \times n}$, where each row represents the weights for each token during feature mixing.
This operation enables the model to focus on different parts of the input for each head through attention weights, thereby capturing various parts of the information from the sequence token representations.
The final output is the current generated token vocabulary conditional probability distribution $\vy \in \R^v$ based on the input instruction representations $\mX_{I}$, image representations $\mX_{V}$ and history generated token representations $\mX_{H}$, where $v$ is the size of the vocabulary. This process can be formatted as:
\begin{equation}
\begin{aligned}
    \vy &\sim p_{\text{model}}(\vy \mid \mathbf{X}_I, \mathbf{X}_V, \mathbf{X}_H), \\
    &\propto \operatorname{softmax} \left(\mathrm{logit}_{\text{model}}(\vy \mid \mathbf{X}_I, \mathbf{X}_V, \mathbf{X}_H)\right),
\end{aligned}
\end{equation}
which calculates the distribution of one token and iterates for the entire response. The sequence generation continues until an EOS (End of Sentence) token is produced, marking the end of the generation and resulting in a complete response.

\vspace{-0.3cm}
\section{Method}\label{sec:method}
\vspace{-0.2cm}

At the core of our method is a solution for image neglect and text inertia, both of which are fundamentally interconnected.
Essentially, as paying more attention to the image, there is a corresponding reduction in the reliance on language priors.
Intuitively, in a conversation centered around an image, the model should devote more attention to the image, thereby allowing it to have a significant impact on the response.
As such, we identify the self-attention map in the token-level generation and augment the image attention in its original directions.
This strategy promotes a more image-centric latent representations.
Additionally, to further mitigate the influence of text inertia, we devide the logits distribution of pure text input into the model's output.

\subsection{Pay More Attention to Image}

\textbf{Extracting the Self-Attention Matrix.} 
We start from a token-level perspective.
The response process in LVLMs is fundamentally generated token by token. 
Each token is generated based on the input image, instruction, and the historically generated response. 
This process is facilitated through a multi-layer attention decoder architecture. 
Consequently, this results in a probability distribution of the vocabulary for the currently generated token.
Our goal is to extract the attention matrix of each attention head at every layer, indicating the influence of each content during inference.

When generating the $k$-th token in the sequence, the input representation for the attention head in the forward process includes the instruction representation $\mX_I=[\vx_{i_1}, \ldots, \vx_{i_{n_I}}]$, image representation $\mX_V = [\vx_{v_1}, \ldots, \vx_{v_{n_V}}]$, and the representation of the historically generated response $\mX_H = [\vx_{h_1}, \ldots, \vx_{h_{n_H}}]$. 
Notably, the image representation considered here is the one that has been processed by the projector.
Essentially, the hidden states of each input layer are $\mX = \text{concat}(\mX_I[1:m], \mX_V, \mX_I[m+1:n_I], \mX_H)$, where the notation $\mX_I[1:m]$ indicates the first $m$ elements in the instruction representation.
Each attention head assigns different degrees of attention to each element during the current token representation decoding process. Our aim is to enhance the attention paid to the image. Therefore, we extract the attention weight values related to the image token for the current generated token, intervene, and then redistribute the attention values of each element through softmax.

\noindent\textbf{Excite model in a trustful direction.} 
There have been attempts in some LLMs works to make the answers generated by LLMs more trustworthy by means of intervention~\cite{iti,ccs,sh2,Azaria_Mitchell_2023,turner2023activation}. 
The implemented approach typically involves intervening with the hidden states. 
As for defining what constitutes a more trustworthy direction, it usually requires additional projection and training to probe this trustful direction. 
In our case, a response that is more image-based is considered more trustworthy. 
Since LVLMs have undergone alignment training, the original attention values provide a direction based on image content. 
As illustrated in \cref{fig:framework}, by amplifying the attention values of image tokens based on the original attention values, we can enhance the trustworthiness of our results.

Another nuance involves our avoidance of choosing the attention head used to shift intervention. 
In the ITI method~\cite{iti}, it is stated that not all attention heads should be subjected to intervention. 
Therefore, they introduce a trustful score to rank each head across all layers and select the top-k heads for intervention. 
In our case, the less trustworthy heads with lower attention values receive less intervention. 
We first extract the attention weights of the image tokens for the 
current generated token from the attention weights $\tilde{\mA}$ before softmax opeartion. We then use the hyper-parameter $\alpha$ to control the step size for intervention. From a single attention head perspective, our method can be expressed as follows: 
\begin{equation}
    \tilde{\mA}_{n, j} = \tilde{\mA}_{n, j} + \alpha \cdot |\tilde{\mA}_{n, j}| \quad \text{for } j = m+1 \text{ to } m+n_V.
\end{equation}
The model's final vocabulary probability distribution is derived from the projection of the hidden states of the last token in the sequence. 
Therefore, we extract the attention weights of the last token $n$ on the image tokens by indexing $\tilde{\mA}_{n, j}$. Following the intervention, we use the softmax function to redistribute the attention values of each token during the reassignment of encoded hidden states.
This procedure is repeated for each subsequent token prediction in an autoregressive manner and is independent of the choice of the decoding algorithm.

\begin{figure*}[tb]    
  \centering    
  \begin{subfigure}{0.3\linewidth}    
    \includegraphics[page=1,width=\linewidth]{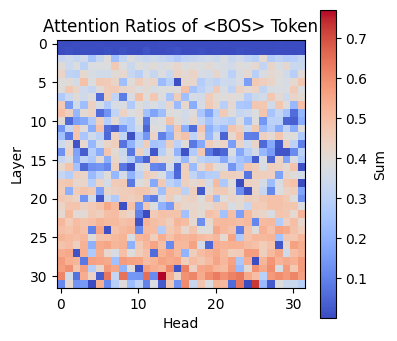}    
    \caption{LLAVA}    
    \label{fig:ratio_llava}    
  \end{subfigure}    
  \begin{subfigure}{0.3\linewidth}    
    \includegraphics[page=2,width=\linewidth]{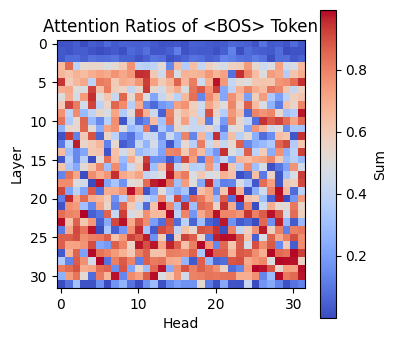}    
    \caption{Minigpt4}    
    \label{fig:ratio_minigpt4}    
  \end{subfigure}
  \begin{subfigure}{0.3\linewidth}    
    \includegraphics[page=2,width=\linewidth]{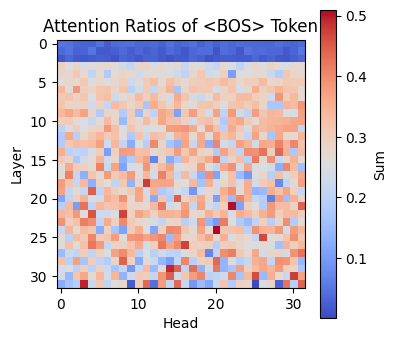}    
    \caption{Shikra}    
    \label{fig:ratio_shikra}    
  \end{subfigure} 
    \vspace{-0.2cm}
  \caption{The BOS token attention ratios of three model. We calculate the attention weights of the BOS token for each head in each layer and display them in a heat map.} 
  \vspace{-0.3cm}
  \label{fig:attn_layer}    
\end{figure*}    

\noindent\textbf{Excite more precisely with an attention mode prior.} The presence of the BOS token, an attention sink pattern~\cite{attn_sink}, in a sentence results in higher attention values during the attention computation process, which may seem counterintuitive. 
The BOS token typically signifies the start of a sentence and, as such, does not carry significant semantic content. 
However, the generation of tokens is significantly influenced by this particular token, a similar pattern that also manifests itself in visual models~\cite{outlier}. 
As mentioned in StreamLLM~\cite{attn_sink}, the pattern of the attention sink emerges when redundant attention values are present.

Naturally, one might infer that when the sink pattern appears, we excite the image token. 
To further investigate this phenomenon, as depicted in \cref{fig:attn_layer}, we find that the sink phenomenon is not overtly evident in the shallow layers. This is because the shallow layers tend to focus more on encoding semantically rich information~\cite{label}. 
When the encoding of semantically rich tokens stabilizes, the attention sink phenomenon arises. 
Therefore, we build upon the judgement of intervention timing by calculating the similarity of the hidden states.

\subsection{Image-Centric Logit Refine}
In \cref{fig:teaser}, we observe a peculiar phenomenon where LVLMs continue to generate identical hallucinated text even when the image is removed from the input. 
This observation naturally leads us to the concept of using the output distribution (when no image is 
in input) as a reference to penalize our initial prediction distribution. 
Therefore, we update the distribution of the generated token by:
\begin{equation}
    \begin{aligned}
        \pmodel &= \gamma \cdot \pmodel(\vy | \mX_V, \mX_I, \mX_H) \\ &\quad - (\gamma - 1) \cdot \pmodel(\vy | \mX_I, \mX_H).       
    \end{aligned}
\end{equation}
This equation effectively reduces the predicted probability based on text alone. The weight $\gamma$ is used to control the degree of penalty applied to the initial prediction distribution. 

This operation is conceptually similar to LLM-CFG~\cite{llm-cfg}. 
Essentially, it provides a guided generation mechanism that allows the model to make informed choices between outputs based on image content and those based on language logic. 
This way, the model can better balance the influence of visual and textual information in its outputs, leading to more contextually accurate and relevant results.

%% file: sections/4.exp.tex
\section{Experiments}\label{sec:exp}
\subsection{Setup}
\noindent\textbf{Baselines.} 
We evaluate the effectiveness of our method on three different models. 
To better compare the impact of image feature tokens after different projectors on our method, we selected two models that use linear projectors, LLAVA and Shikra, as well as one model that uses resamplers, Minigpt4. 
Additionally, for a more convincing comparison, we report on three decoding methods for comparison: greedy, beam search, and nucleus sample. 
We also selected the OPERA~\cite{opera} method, which is an improvement on beam search, and the VCD~\cite{vcd} method, which is an improvement on nucleus sampling, to compare with our results. We used the default hyperparameters from the open-source versions of these two methods.

\noindent\textbf{Implementation Details.} 
As different models have different lengths of image tokens, leading to different degrees of image neglect, to better align with the image sequence length of the model, we set $\alpha=0.5$ for LLAVA, $\alpha=0.6$ for Shikra with long image token sequence lengths and $\alpha=0.2$ for resampler models with short image token sequences. 
As text inertia is independent of the image token length, we continuously use $\gamma=1.1$. 
Apart from this, in the beam search tests, the beam number is set to 5 for all methods, and in the nucleus sample tests, all the common parameters are consistent.

\subsection{Benchmark \& Evaluation Metrics}
\textbf{CHAIR~\cite{chair}.} Caption Hallucination Assessment with Image Relevance (CHAIR) is a widely used metric in image captioning tasks. CHAIR operates by creating a set of ground-truth object labels for each image. Any object mentioned in the caption that does not exist in the label set is considered a hallucinated object. CHAIR comprises two evaluation dimensions: instance-level and sentence-level, represented as $\mathrm{CHAIR_I}$ and $\mathrm{CHAIR_S}$, respectively. These are calculated in the following manner:
\begin{gather}  
    \mathrm{CHAIR_I} = \frac{|\{\text{hallucinated objects}\}|}{\text{all mentioned objects}}, \\  
    \mathrm{CHAIR_S} = \frac{|\{\text{captions with hallucinated objects}\}|}{\text{all captions}}.   
\end{gather}   
We conducted experiments based on the validation set of MSCOCO 2014. 
Given ``\texttt{Please help me describe the image in detail.}'' as prompt, we subsequently employ the CHAIR metric to evaluate the generated description.
To evaluate hallucinations in long sequence generation, we adopted the same setup as used in LURE~\cite{lure} and OPERA~\cite{opera}, setting the $\text{max\_new\_tokens}$ parameter to 512 and randomly sampled 500 instances for evaluation.

\noindent\textbf{POPE~\cite{pope}.} The Polling-based Object Probing Evaluation (POPE) is a evaluation metric designed in the VQA paradigm.
POPE serves as a metric for assessing object hallucination, evaluating hallucinations by asking LVLMs questions such as ``\texttt{Is there a <object> in the image?}'' Here, \texttt{<object>} is replaced with the constructed ground-truth object from three different types of splits. 
In the ``random'' split, objects are randomly selected from the entire dataset for evaluation. 
In the ``popular'' split, objects are chosen from those most frequently appearing in the dataset. 
In the ``adversarial'' split, objects that are highly related to the image objects are selected for evaluation. 
We conduct our evaluation on the COCO dataset with 500 images, with each image having 6 questions for each split of POPE. 
We evaluate the performance of the model in object recognition tasks using both accuracy score and F1 score.
Besides, in order to more comprehensively examine hallucinations in multi-turn chats, we have structured the evaluation in both single-turn and multi-turn dialogue forms.

\noindent\textbf{MMHal-Bench~\cite{mmhalbench}.} For further evaluation of our method on some challenging datasets, we choose MMHal-Bench, which is designed with 96 image-question pairs, spread across 8 question categories $\times$ 12 object topics. 
It contains eight types of questions about object attributes, adversarial objects, comparisons, counting, spatial relations, environment, holistic descriptions, and others to comprehensively assess the model’s hallucination performance on high-difficulty datasets. 
Essentially, it is also a VQA-based evaluation, but unlike the existence-based examination in POPE, its questions also include some logical considerations. 
Therefore, for evaluation on MMHal-Bench, we first need to answer questions and then use GPT-4 to score the answer based on the response and the ground-truth answer. 
The evaluation results include the model’s scores across all question categories, and the overall score represents the average of these scores.

\noindent\textbf{GPT-4v Assisted Evaluation.} To further evaluate the model’s performance in image description tasks, we can move beyond the CHAIR metric, which is based on information extraction and only considers object hallucination. 
We can use GPT-4v for open evaluation. As with previous evaluations~\cite{opera,vcd}, we sample 50 images on the COCO dataset for evaluation. 
We construct prompts, input images into GPT-4v, along with the descriptions responses from two assistants. 
GPT-4v evaluation takes into account two dimensions: Accuracy and Detailedness, denoted respectively as C and D.
Detailed prompt construction can be found in the appendix.

\begin{table*}[t]            
\centering          
\caption{\textbf{CHAIR hallucination evaluation results} on three LVLMs. $\mathrm{CHAIR}$~\cite{chair} is employed as the evaluation metric, where a smaller number indicates less hallucinations. OPERA is a decoding method based on beam search, VCD is a decoding method based on nucleus sampling, and PAI is an inference intervention method that can be collaborated with any decoding method.} 
\begin{tabular}{@{}llcccccc@{}}           
\toprule          
\multirow{2}{*}{\textbf{Decoding}} & \multirow{2}{*}{\textbf{Method}} & \multicolumn{2}{c}{\textbf{LLAVA~\cite{llava-1.5}}} & \multicolumn{2}{c}{\textbf{Minigpt4~\cite{minigpt4}}} & \multicolumn{2}{c}{\textbf{Shikra~\cite{shikra}}}\\    
\cmidrule(lr){3-4} \cmidrule(lr){5-6} \cmidrule(lr){7-8}    
  &  & $\mathrm{CHAIR_S}$ & $\mathrm{CHAIR_I}$ & $\mathrm{CHAIR_S}$ & $\mathrm{CHAIR_I}$ & $\mathrm{CHAIR_S}$ & $\mathrm{CHAIR_I}$ \\            
\midrule            
\multirow{2}{*}{\makecell[l]{Greedy}} & Vanilla & 46.6 & 13.4 & 32.8 & 11.1 & 51.2 & 14.4\\            
 & PAI & \textbf{24.8} & \textbf{6.9} & \textbf{26.3} & \textbf{8.8} & \textbf{37.6} & \textbf{10.0}\\        
\cmidrule(l){1-8}      
\multirow{3}{*}{\makecell[l]{Beam Search}} & Vanilla & 46.4 & 14.3 & 46.6 & 13.4 & 53.0 & 14.7\\            
 & OPERA~\cite{opera} & 44.6 & 14.4 & 30.1 & 9.8 & 36.8 & 12.4\\           
 & PAI & \textbf{21.8} & \textbf{5.6} & \textbf{24.8} & \textbf{6.9} & \textbf{35.8} & \textbf{11.4}\\          
\cmidrule(l){1-8}      
\multirow{3}{*}{\makecell[l]{Nucleus}} & Vanilla & 58.2 & 18.2 & 32.7 & 11.9 & 57.9 & 16.4\\            
 & VCD~\cite{vcd} & 51.8 & 15.1 & 34.8 & 11.5 & 57.6 & 16.3\\           
 & PAI & \textbf{43.4} & \textbf{14.7} & \textbf{26.7} & \textbf{10.3} & \textbf{49.9} & \textbf{13.2}\\         
\bottomrule          
\end{tabular}    
\label{tab:chair}            
\end{table*}  

\subsection{Experimental Results}
In this section, we analyze the performance of PAI across various hallucination evaluation tasks, including long image description, simplified VQA answer, construction of metric evaluation, and leveraging the near-human cognitive capabilities of GPT-4/GPT-4v as evaluation methods. 
For further analysis, please refer to the appendix.

\begin{table*}[t]            
\centering        
\caption{\textbf{Quantitative comparison} on POPE. The best results are in \textbf{bold}.} 
\resizebox{\linewidth}{!}{
\begin{tabular}{@{}lccccccccccccc@{}}            
\toprule          
\multirow{3}{*}{\textbf{Decoding}} & \multirow{3}{*}{\textbf{Method}} & \multicolumn{4}{c}{\textbf{LLAVA~\cite{llava-1.5}}} & \multicolumn{4}{c}{\textbf{Minigpt4~\cite{minigpt4}}} & \multicolumn{4}{c}{\textbf{Shikra~\cite{shikra}}} \\           
\cmidrule(r){3-4} \cmidrule(l){5-6} \cmidrule(l){7-8} \cmidrule(l){9-10} \cmidrule(l){11-12} \cmidrule(l){13-14}
 & & \multicolumn{2}{c}{\textbf{Single-turn}} & \multicolumn{2}{c}{\textbf{Multi-turn}} & \multicolumn{2}{c}{\textbf{Single-turn}} & \multicolumn{2}{c}{\textbf{Multi-turn}} & \multicolumn{2}{c}{\textbf{Single-turn}} & \multicolumn{2}{c}{\textbf{Multi-turn}} \\        
 & & {Acc} & {F1} & {Acc} & {F1} & {Acc} & {F1} & {Acc} & {F1} & {Acc} & {F1} & {Acc} & {F1} \\        
\midrule            
\multirow{2}{*}{Greedy} & Vanilla & 84.76 & 85.51 & 85.69 & 84.27 & 74.39 & 73.78 & 78.0 & 77.77 & 81.56 & \textbf{82.05} & 78.16 & 78.77 \\            
 & PAI & \textbf{85.82} & \textbf{85.97} & \textbf{87.54} & \textbf{86.83} & \textbf{75.2} & \textbf{76.2} & \textbf{80.82} & \textbf{81.6} & \textbf{82.3} & 81.47 & \textbf{78.57} & \textbf{79.11} \\        
\cmidrule{1-14}   
\multirow{2}{*}{Beam Search} & Vanilla & 84.9 & 84.9 & 83.57 & 81.13 & 72.54 & 67.73 & 71.78 & 67.92 & 81.52 & \textbf{81.89} & 78.18 & 78.13 \\       
 & PAI & \textbf{86.33} & \textbf{85.89} & \textbf{86.42} & \textbf{85.27} & \textbf{73.22} & \textbf{69.36} & \textbf{74.15} & \textbf{71.72} & \textbf{82.25} & 81.01 & \textbf{80.72} & \textbf{80.02} \\        
\cmidrule{1-14}   
\multirow{3}{*}{Nucleus} & Vanilla & 80.25 & 81.32 & 82.87 & 81.63 & 57.52 & 58.16 & 64.49 & 63.89 & 79.55 & 80.28 & 76.54 & 77.12 \\   
 & VCD~\cite{vcd} & 79.53 & 81.02 & 84.03 & 83.3 & \textbf{59.97} & 58.7 & 69.72 & 69.38 & 80.22 & 80.5 & 76.38 & 77.07 \\        
 & PAI & \textbf{81.72} & \textbf{82.87} & \textbf{85.03} & \textbf{83.97} & 59.19 & \textbf{60.8} & \textbf{70.24} & \textbf{70.27} & \textbf{81.49} & \textbf{80.71} & \textbf{77.99} & \textbf{78.47} \\       
\bottomrule          
\end{tabular}  
}
\label{tab:pope}            
\end{table*}

\noindent\textbf{Results on long sequence hallucination evaluation. }
The experimental results are presented in \cref{tab:chair}. 
As our approach is an inference intervention method, it differs from previous decoding hallucination mitigation methods that primarily concentrate on improving a single decoding method.
We have tested our method on three decoding techniques. 
Our method has achieved hallucination mitigation on all three decoding methods used by the three models. 
However, when integrated with the nucleus, a sampling-based method, the hallucination reduction brought about by our method is not significant. 
This may be because even though our method has increased the priority of trustful tokens, the sample set during nucleus decoding still contains many hallucination tokens.

Moreover, while OPERA significantly mitigates hallucinations, its time efficiency is considerably higher compared to vanilla. 
In contrast, our method not only has almost the same time efficiency as vanilla, but it also performs better in reducing hallucination issues. 
Compared to VCD, during the generation process of long sequence tasks, the introduction of visual uncertainty during decoding sometimes leads to more hallucination descriptions. 
However, our method can reduce the proportion of hallucination words in the sample pool.

\noindent\textbf{Results on single-turn and multi-turn hallucination evaluation.} Unlike the CHAIR evaluation, POPE is in a VQA format, so the response is brief, answering only ``Yes'' or ``No''. The phenomena of text inertia and image neglect may not be as noticeable under this setting, especially for single-turn. However, our method still achieved a notable improvement compared to the vanilla decoding method in single-turn. 
As for multi-turn evaluation, which involves a longer context, our method achieves more significant improvement as presented in~\cref{tab:pope}. 

\begin{figure*}[t]  
  \centering    
  \begin{subfigure}{0.32\linewidth}    
    \includegraphics[page=1,width=\linewidth]{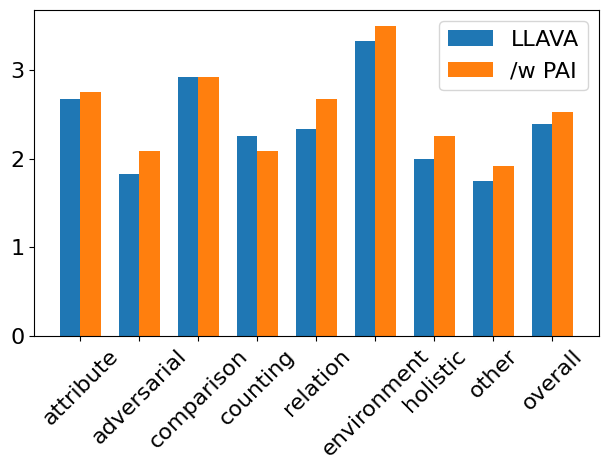}    
    \caption{LLAVA}    
    \label{fig:mmhal_llava}    
  \end{subfigure}    
  \begin{subfigure}{0.32\linewidth}    
    \includegraphics[page=2,width=\linewidth]{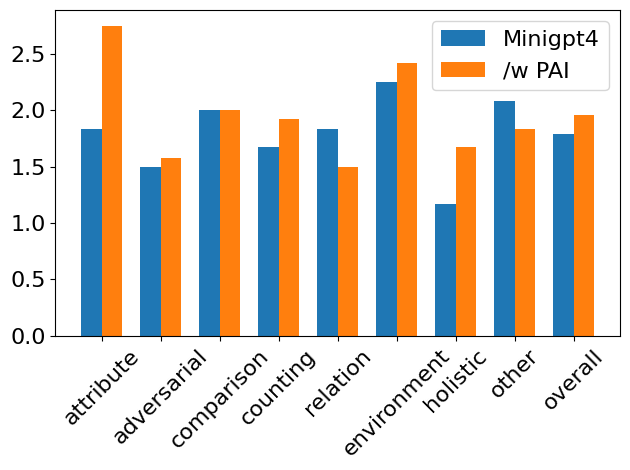}    
    \caption{Minigpt4}    
    \label{fig:mmhal_minigpt4}    
  \end{subfigure}
  \begin{subfigure}{0.32\linewidth}    
    \includegraphics[page=2,width=\linewidth]{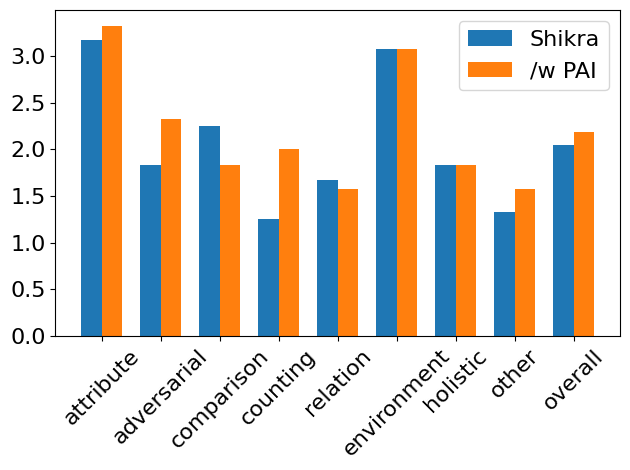}    
    \caption{Shikra}    
    \label{fig:mmhal_shikra}    
  \end{subfigure}
  \caption{\textbf{Quantitative comparison} on the MMHal-Bench. Higher scores indicate better performance.}    
  \label{fig:mmhal}    
\end{figure*}       

\noindent\textbf{Results on hallucination evaluation in comprehensive general scenarios.} 
The experimental results, as shown in \cref{fig:mmhal}, indicate that for some more image-based question types, such as object attributes, adversarial objects, and holistic questions, the answers are more accurate when inference intervention with PAI is applied, and there is a certain degree of improvement across all models. However, for some logical questions, such as comparisons and relations, there is no noticeable improvement after intervention. In summary, through the overall metric, i.e., the average of the eight evaluation dimensions, there is a certain degree of improvement compared to the baseline after incorporating PAI.

\noindent\textbf{Results on human-like GPT-4v assisted hallucination evaluation.} 
The experimental results, as shown in \cref{tab:gpt4v}, indicate that even when more comprehensive hallucination evaluation dimensions are added, our method, compared to the greedy decoding method, can provide more accurate responses on all three models without losing detail in the description. Given that GPT-4v’s visual understanding and language logic capabilities have reached a level close to that of humans, it can more comprehensively illustrate the performance improvements brought about by our method.
\begin{table}     
    \centering    
    \caption{\textbf{Results on GPT-4V evaluation.} The best results are in \textbf{bold}.}                    
    \begin{tabular}{@{}lcccccc}                    
    \toprule                  
    \multirow{2}{*}{\textbf{Method}} & \multicolumn{2}{c}{\textbf{LLAVA}} & \multicolumn{2}{c}{\textbf{Minigpt4}} & \multicolumn{2}{c}{\textbf{Shikra}} \\                   
    \cmidrule(r){2-3} \cmidrule(lr){4-5} \cmidrule(l){6-7}                
     & {C} & {D} & {C} & {D} & {C} & {D} \\                
    \midrule                    
    Greedy & 5.62 & 5.24 & 5.8 & 5.74 & 5.54 & \textbf{5.25}\\                    
    PAI & \textbf{6.46} & \textbf{5.36} & \textbf{7.04} & \textbf{5.89} & \textbf{6.04} & 5.05\\              
    \bottomrule                  
    \end{tabular}                   
    \label{tab:gpt4v}                    
\end{table}

\begin{table*}[t!]  
    \centering  
    \caption{\textbf{Ablation Study of the Hyperparameter $\alpha$}. When $\alpha$ becomes excessively large, resulting in an unbalanced response, we terminate the experiment and denote this with a dash (-). F1 values that are considered outliers are highlighted in \textcolor{red}{red}.}  
    \label{tab:abla_alpha}  
    \begin{tabular}{@{}lccccccccccc}                      
        \toprule                    
        \multirow{2}{*}{\textbf{$\alpha$}} & \multirow{2}{*}{\textbf{$\gamma$}} & \multirow{2}{*}{L} & \multicolumn{3}{c}{\textbf{LLaVA}} & \multicolumn{3}{c}{\textbf{Minigpt4}} & \multicolumn{3}{c}{\textbf{Shikra}} \\                      
        \cmidrule(r){4-6} \cmidrule(lr){7-9} \cmidrule(lr){10-12}           
         & & & {$\mathrm{CHAIR_S}$} & {$\mathrm{CHAIR_I}$} & {F1} & {$\mathrm{CHAIR_S}$} & {$\mathrm{CHAIR_I}$} & {F1} & {$\mathrm{CHAIR_S}$} & {$\mathrm{CHAIR_I}$} & {F1}\\                      
        \midrule                      
        - & - & - & 46.2 & 13.8 & 75.9 & 31.6 & 10.5 & 69.4  & 56.2 & 15.8 & 74.6\\      
        \midrule  
        0.1 & 1.1 & $\checkmark$ & 47.4 & 13.8 & 76.9 & 25.0 & 8.0 & 69.9  & 56.4 & 15.0 & 75.7\\           
        0.2 & 1.1 & $\checkmark$ & 47.4 & 13.6 & 76.9 & \textbf{21.2} & \textbf{7.7} & 70.3 & 57.2 & 15.6 & 75.2\\       
        0.4 & 1.1 & $\checkmark$ & 42.4 & 12.4 & 76.9 & 15.1 & 6.3 & 67.4  & 55.2 & 14.4 & 75.3\\       
        0.5 & 1.1 & $\checkmark$ & \textbf{24.6} & \textbf{6.8} & 74.7 & 5.4 & 2.9 & \textcolor{red}{61.5} & 51.6 & 13.2 & 75.8\\        
        0.6 & 1.1 & $\checkmark$ & 7.8 & 4.8 & \textcolor{red}{62.6} & - & - & - & \textbf{37.2} & \textbf{10.6} & 76.7 \\      
        0.7 & 1.1 & $\checkmark$ & - & - & - & - & - & - & 14.4 & 6.1 & \textcolor{red}{64.9} \\     
        \bottomrule                    
    \end{tabular}    
\end{table*}  

% \begin{figure*}[t]          
%   \centering        
%   \begin{minipage}[t]{0.48\linewidth}      
%     \centering      
%     \begin{subfigure}[t]{\linewidth}          
%       \includegraphics[clip, trim={0cm 0cm 0cm 1cm}, page=1,width=\linewidth]{./images/case_2.pdf}          
%     \end{subfigure}          
%   \end{minipage}      
%   \hfill      
%   \begin{minipage}[t]{0.48\linewidth}      
%     \centering      
%     \begin{subfigure}[t]{\linewidth}          
%       \includegraphics[clip, trim={0cm 0cm 0cm 1cm}, page=2,width=\linewidth]{./images/case_2.pdf}          
%     \end{subfigure}          
%   \end{minipage}      
%   % \vspace{}
%   \begin{minipage}[t]{0.48\linewidth}      
%     \centering      
%     \begin{subfigure}[t]{\linewidth}          
%       \includegraphics[clip, trim={0cm 1cm 0cm 1cm}, page=3,width=\linewidth]{./images/case_2.pdf}          
%     \end{subfigure} 
%   \end{minipage}      
%   \hfill      
%   \begin{minipage}[t]{0.48\linewidth}      
%     \centering      
%     \begin{subfigure}[t]{\linewidth}          
%       \includegraphics[clip, trim={0cm 1cm 0cm 1cm}, page=4,width=\linewidth]{./images/case_2.pdf}          
%     \end{subfigure}        
%   \end{minipage}      
%   \caption{Illustration of hallucination correction by our proposed PAI with two samples. The left panel shows the response from LLAVA, while the right panel presents the response from LLAVA when integrated with PAI. The descriptions containing hallucinations are emphasized in \textcolor{red}{red}.}   
%   \vspace{-0.5cm}
%   \label{fig:case2}     
% \end{figure*}   

\subsection{Ablation Study}
Our method, PAI, consists of two stages of interventions. In the first stage, during forward inference, the hyperparameter $\alpha$ is utilized to set the scale of intervention. Simultaneously, the layer prior, represented as ``L'', is used to determine the attention layer for intervention. The second stage unfolds during the decoding process, where we mitigate text inertia by subtracting the logits distribution that results from inputs devoid of image information. In this stage, the scale is managed by the parameter $\gamma$.

We use LLaVA-1.5 as the representative LVLM baseline and the greedy decoding method as the basic baseline to compare the impact of our hyperparameters on the task of long sequence image description. To evaluate our method, we choose the CHAIR metric. However, since CHAIR only evaluates the hallucination problem, we have incorporated the F1 score to consider information richness and accuracy. This makes the comparison with the CHAIR metric fairer when the F1 scores are similar. In the above, the F1 scores of the various methods differ slightly, so we have not included this somewhat redundant measure. The F1 score is calculated by counting the objects included in the description, the objects in the ground-truth sets, and the hallucinated objects.

\noindent\textbf{Effects of $\alpha$ in Exciting Image Attention.}
In the process of exciting the attention values of image tokens, we introduce a parameter $\alpha$ to control the amplification scale. As shown in ~\cref{tab:abla_alpha}, different LVLMs exhibit varying sensitivity to the amplification scale. This sensitivity not only depends on the length of the model's image tokens (e.g., the image token length of LLaVA-1.5 is 576, while that of Minigpt4 is only 32), but also the original attention weights distribution. 

However, a commonality across these models is that an appropriate amplification scale can achieve a balance between the number of hallucinated objects in the description and the amount of information conveyed. If the scale is too small, the description may still contain many hallucinated objects. Conversely, if the scale is too large, the amount of information in the response will decrease.

\begin{table*}[t]  
\centering      
\caption{\textbf{Ablation Study of Hyperparameter $\gamma$ and Layer Prior ``L''.} Results are presented for the models LLAVA (left), Minigpt4 (middle), and Shikra (right).}  
\label{tab:abla_2}  
\small % Or \footnotesize if you want it even smaller  
\resizebox{.33\linewidth}{!}{  
    \begin{tabular}{@{}lccccccccccc}                                
        \toprule                              
        $\alpha$ & $\gamma$ & L & {$\mathrm{CHAIR_S}$} & {$\mathrm{CHAIR_I}$} & {F1} \\                            
        \midrule                                
        - & - & - & 46.2 & 13.8 & 75.9 \\                
        \midrule              
        0.5 & 1.1 & $\checkmark$ & 24.6 & 6.8 & 75.7 \\                    
        0.5 & 1.2 & $\checkmark$ & \textbf{23.4} & \textbf{6.6} & 75.7 \\                 
        0.5 & 1.3 & $\checkmark$ & 24.6 & 7.5 & 74.4 \\                 
        0.5 & 1.5 & $\checkmark$ & 25.0 & 9.6 & 74.6 \\                  
        0.5 & 2.0 & $\checkmark$ & 24.2 & 7.1 & 74.0 \\               
        \midrule              
        0.5 & 1.1 & $\checkmark$ & \textbf{24.6} & \textbf{6.8} & 75.7 \\             
        0.5 & 1.1 & $\times$ & 20.6 & 6.7 & \textcolor{red}{71.7} \\             
        \bottomrule                              
    \end{tabular} 
}%  
\resizebox{.33\linewidth}{!}{ 
    \begin{tabular}{@{}lccccccccccc}                                
        \toprule                              
        $\alpha$ & $\gamma$ & L & {$\mathrm{CHAIR_S}$} & {$\mathrm{CHAIR_I}$} & {F1} \\                            
        \midrule                                
        - & - & - & 31.6 & 10.5 & 69.4 \\                
        \midrule              
        0.2 & 1.1 & $\checkmark$ & 21.2 & 7.7 & 70.3 \\                    
        0.2 & 1.2 & $\checkmark$ & 16.4 & \textbf{6.2} & 70.3 \\                 
        0.2 & 1.3 & $\checkmark$ & \textbf{14.4} & 6.9 & 69.5 \\                 
        0.2 & 1.5 & $\checkmark$ & 12.6 & 5.4 & \textcolor{red}{67.1} \\                  
        0.2 & 2.0 & $\checkmark$ & 8.4 & 4.6 & \textcolor{red}{64.3} \\               
        \midrule              
        0.2 & 1.1 & $\checkmark$ & \textbf{21.2} & \textbf{7.7} & 70.3 \\             
        0.2 & 1.1 & $\times$ & 22.3 & 8.4 & 70.3 \\             
        \bottomrule                              
    \end{tabular}
}%  
\resizebox{.33\linewidth}{!}{   
    \begin{tabular}{@{}lccccccccccc}                                
        \toprule                              
        $\alpha$ & $\gamma$ & L & {$\mathrm{CHAIR_S}$} & {$\mathrm{CHAIR_I}$} & {F1} \\                            
        \midrule                                
        - & - & - & 56.2 & 15.8 & 74.6 \\                
        \midrule              
        0.6 & 1.1 & $\checkmark$ & \textbf{37.2} & 10.6 & 76.7 \\                    
        0.6 & 1.2 & $\checkmark$ & 40.2 & 10.5 & 76.0 \\                 
        0.6 & 1.3 & $\checkmark$ & 40.6 & 10.5 & 76.0 \\                 
        0.6 & 1.5 & $\checkmark$ & 40.2 & 10.5 & 75.3 \\                  
        0.6 & 2.0 & $\checkmark$ & 39.4 & \textbf{10.1} & 76.2 \\               
        \midrule              
        0.6 & 1.1 & $\checkmark$ & \textbf{37.2} & \textbf{10.6} & 76.7 \\             
        0.6 & 1.1 & $\times$ & 56.4 & 15.7 & 74.9 \\             
        \bottomrule                              
    \end{tabular}
} 
\end{table*}

\noindent\textbf{Effects of Layer Prior in Exciting Image Attention.} 
We further investigate the control of the intervention layer for exciting image attention. As observed in \cref{tab:abla_2}, the introduction of this prior does indeed improve the performance of our method. However, when there is no control over the intervention layer and interventions are applied to all layers, the performances of different models exhibit some variations. For Shikra, in the absence of the layer prior, our method's results regress to the baseline. For both LLaVA and Minigpt4, the loss of the layer prior causes some unfavorable fluctuations in both the CHAIR metric and the F1 score.

\noindent\textbf{Effects of $\gamma$ in Mitigating Language Prior.} 
\cref{tab:abla_2} presents the results of an ablation study focusing on $\gamma$, which adjusts the balance between output distributions from conditioned inputs with excited image tokens and pure text inputs. Unlike the other two models, the Minigpt4 model is highly sensitive to $\gamma$. When $\gamma$ is too large, it can lead to uncontrollable model behavior. The experimental results suggest that maintaining $\gamma$ within a relatively small range, such as 1.1 - 1.2, yields the most stable performance.

%% file: sections/5.conclusion.tex
\section{Conclusion and Limitation}\label{sec:conclusion}
In this paper, we first analyze the causes and manifestations of hallucinations in LVLMs. We propose a phenomenon termed "text inertia," wherein the model continues to produce the same hallucinatory descriptions even when no image input is provided. This issue fundamentally stems from the model's neglect of image tokens. Consequently, we introduce the PAI method to intervene in the model's inference process, steering it towards an image-based and trustworthy direction. This is a training-free method and does not require any external tools. Extensive experiments on multiple benchmarks and LVLMs have validated the effectiveness of PAI in mitigating hallucination issues.

\noindent\textbf{Limitations:} (1) The language decoders of existing open-source LVLMs are primarily models from the LLaMA-family. It is worth exploring whether the issues of image neglect and text inertia are introduced by LLaMA. (2) As we described, our method fundamentally alleviates the image neglect issue during model inference. Its upper limit depends on the capabilities of the well-trained model. Therefore, it is worth investigating whether incorporating this issue as a loss during the training process could lead to further performance improvements.

%% file: sections/6.ref.tex
\bibliographystyle{splncs04}
\bibliography{main.bib}

%% file: sections/7.appendix.tex
\clearpage
\appendix
\renewcommand\thesection{\Alph{section}}
\renewcommand\thefigure{S\arabic{figure}}
\renewcommand\thetable{S\arabic{table}}
\renewcommand\theequation{S\arabic{equation}}
\setcounter{figure}{0}
\setcounter{table}{0}
\setcounter{equation}{0}

\section*{Appendix}

\section{Text Inertia Detection Process}
\label{sec:sup_a}
To detect the text inertia phenomenon in LVLMs during image description tasks, we approach this issue in two stages. In the first stage, we employ the CHAIR metric to identify the indices of hallucinated objects in the LVLMs’ descriptions. In the following stage, we extract the tokens preceding the hallucinated objects, feed them into the LVLMs, and proceed with generation, excluding the image from the input. In other words, we only provide the LVLMs with the task instruction and history response up to the index of the hallucinated object. Finally, we extract the first ten tokens from the newly generated output and use GPT-4 to determine whether the same object has been generated. Our prompt structure is shown in ~\cref{tab:text_inertia_det}.

\begin{table}[ht]    
\caption{The Prompt Used for Text Inertia Detection.}    
\label{tab:text_inertia_det}    
\centering    
\begin{tabular}{p{0.95\linewidth}}    
\hline    
\textbf{GPT-4 Prompt} \\      
\hline      
Please assist me in determining whether the following descriptions include the specified object. Simply respond with ``Yes'' or ``No''. Consider synonyms and similar expressions. \par  
\\ \par  
[Object] \{object\}  
\\ \par  
[Description] \{description\}  
\\ \par  
\hline      
\end{tabular}    
\end{table}    

Here, \{object\} is replaced by the hallucinated object that we extract each time, and \{description\} is replaced by the first ten tokens of the newly generated conditioned description. When the model answers ``Yes'', it indicates the presence of the text inertia phenomenon.

\begin{table}[ht]  
    \centering  
    \caption{Performance of Our Method PAI on the LLaVA-1.5 (13B) Model. We present the average Accuracy and F1 score across the three splits of POPE.}  
    \label{tab:comparison_scale}  
        \begin{tabular}{@{}llcccccc}  
            \toprule  
            \multirow{2}{*}{\textbf{Model}} & \multirow{2}{*}{\textbf{Method}} & \multicolumn{3}{c}{\textbf{CHAIR}} & \multicolumn{2}{c}{\textbf{POPE}} \\  
            \cmidrule(r){3-5} \cmidrule(lr){6-7}  
            & & {$\mathrm{CHAIR_S}$} & {$\mathrm{CHAIR_I}$} & {F1} & {Acc} & {F1} \\  
            \midrule  
            \multirow{2}{*}{LLaVA-1.5(13B)} & Greedy & 44.0 & 12.7 & 77.3 & 85.47 & 86.60 \\  
            & PAI & \textbf{33.0} & \textbf{9.2} & 77.8 & \textbf{86.82} & \textbf{87.80} \\  
            \bottomrule  
        \end{tabular}  
\end{table}  
  
\begin{table}[ht]  
    \centering  
    \caption{Comparison with Different Intervention Method. We present the average Accuracy and F1 score across the three splits of POPE.}  
    \label{tab:comparison_method}  
        \begin{tabular}{@{}llcccccc}  
            \toprule  
            \multirow{2}{*}{\textbf{Model}} & \multirow{2}{*}{\textbf{Method}} & \multicolumn{3}{c}{\textbf{CHAIR}} & \multicolumn{2}{c}{\textbf{POPE}} \\  
            \cmidrule(r){3-5} \cmidrule(lr){6-7}  
            & & {$\mathrm{CHAIR_S}$} & {$\mathrm{CHAIR_I}$} & {F1} & {Acc} & {F1} \\  
            \midrule  
            \multirow{3}{*}{LLaVA-1.5(7B)} & Greedy & 46.2 & 13.8 & 75.9 & 84.76 & 85.59 \\  
            & PH & 52.8 & 14.2 & 76.9 & 84.11 & 85.20 \\  
            & PAI & \textbf{23.4} & \textbf{6.6} & 75.7 & \textbf{86.13} & \textbf{86.42} \\  
            \midrule  
            \multirow{3}{*}{LLaVA-1.5(13B)} & Greedy & 44.0 & 12.7 & 77.3 & 85.47 & 86.60 \\  
            & PH & 46.6 & 12.8 & 77.3 & 85.69 & 86.78 \\  
            & PAI & \textbf{33.0} & \textbf{9.2} & 77.8 & \textbf{86.82} & \textbf{87.80} \\  
            \bottomrule  
        \end{tabular}  
\end{table}  

\section{What if Model Scales Up?}
We performed experiments on the LLaVA-1.5-13B scale model to assess how the performance of PAI varies with increasing model scale. As shown in ~\cref{tab:comparison_scale}, our method maintains its robustness regardless of the scale of the model. Notably, it continues to alleviate the hallucination issue in both long sequence and short VQA tasks as the model scale increases.

\section{Comparative Experimental Results}
We compare our method with a different intervention approach during model inference. \textbf{Prompt Highlighter}~\cite{prompthighlighter} proposes an interactive technique for constructing visual prompts to steer model generation. Essentially, it adds a value of the same size to the attention mask of the tokens in the user's guidance area to influence the model's output. To automate the intervention process for evaluation, we adopt an approach similar to Prompt Highlighter. This involves adding a constant value of 1.0 to the attention mask of all image tokens, where 1.0 is the default value provided by their open-sourced code.

The final experimental results are presented in~\cref{tab:comparison_method}. The findings indicate that uniformly applying intervention across the entire image is less effective than enhancing image attention based on the original scale.
% \subsubsection{Prompt Highlighter/+ - scale}

% \subsubsection{With opera cd}

% \subsubsection{MMHal specific results}

% \subsubsection{DifferentScale}
\newpage
\begin{figure*}[tb]  
    \centering  
    \includegraphics[width=0.9\linewidth]{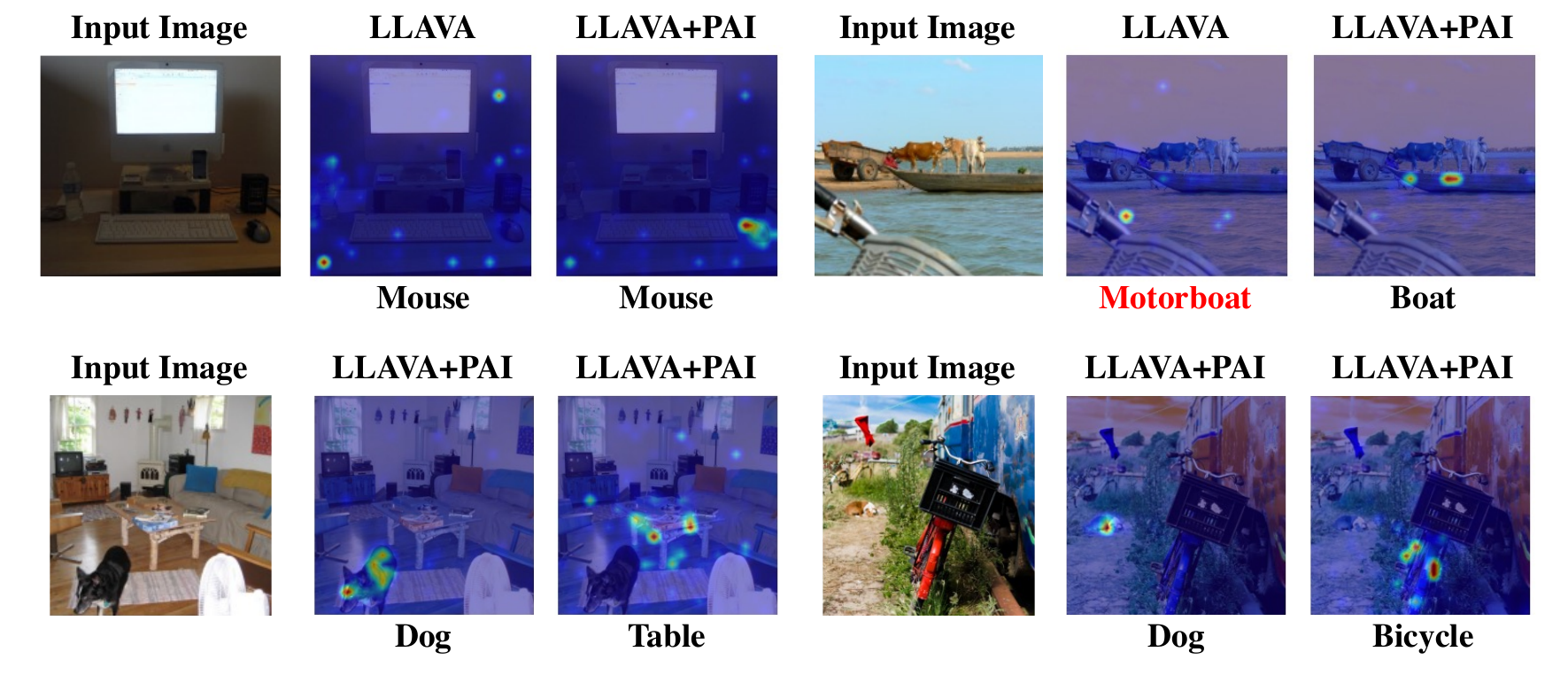}  
    \caption{Visualization of the self-attention maps for each object token with and without PAI over LLAVA. The hallucinated objects are higlighted in \textcolor{red}{red}.}  
    \label{exp}  
\end{figure*} 

\section{Obtaining Explainable LVLMs.} 
The extent to which attention can serve as an explanatory mechanism has been extensively studied~\cite{rollout,tcav}. 
Within the scope of LVLMs, the self-attention maps in the LLaMA decoder have been regarded as a natural explanation for the model. 
However, these explanatory results are only available for LLAVA and Shikra, due to the usage of linear projection that preserves all image patch token features. 
The model, which employs a resampler to derive the explanation, has to reverse-engineer the abstracted image token back to the input image patch token. However, tracing back the image patch after it has been encoded by the visual encoder and resampled into an image token proves to be challenging.

As illustrated in \cref{exp}, the original LLAVA identifies the correct object, a mouse, and hallucinates an object, a motorboat, both of which have limited interpretability as the results do not focus on the main body of the described object. 
With our method of layer intervention, the response becomes more image-centric. 
The description aligns better with the image location, which can all be considered as a more faithful explanation.
\clearpage

\begin{table*}[t]  
\caption{The prompt used for GPT-4V evaluation}  
\label{tab:your_label}  
\centering  
\begin{tabular}{p{\textwidth}}  
\hline  
\textbf{GPT-4V Prompt} \\    
\hline    
You are required to score the performance of two AI assistants in describing a given image. You should pay extra attention to the hallucination, which refers to the part of descriptions that are inconsistent with the image content, such as claiming the existence of something not present in the image or describing incorrectly in terms of the counts, positions, or colors of objects in the image. Please rate the responses of the assistants on a scale of 1 to 10, where a higher score indicates better performance, according to the following criteria: \par  
1: Accuracy: whether the response is accurate with respect to the image content. Responses with fewer hallucinations should be given higher scores. \par  
2: Detailedness: whether the response is rich in necessary details. Note that hallucinated descriptions should not count as necessary details. \par  
Please output the scores for each criterion, containing only two values indicating the scores for Assistant 1 and 2, respectively. The two scores are separated by a space. Following the scores, please provide an explanation of your evaluation, avoiding any potential bias and ensuring that the order in which the responses were presented does not affect your judgment. \par 
 \\ \par
[Assistant 1] \par  
\{Response of Assistant 1\} \par  
[End of Assistant 1] \par 
 \\ \par
[Assistant 2] \par  
\{Response of Assistant 2\} \par  
[End of Assistant 2] \par
 \\ \par
Output format: \par  
Accuracy: \textless Scores of the two answers\textgreater \par  
Reason: \par  
 \\ \par
Detailedness: \textless Scores of the two answers\textgreater \par  
Reason: \par   
 \\ \par
\hline    
\end{tabular}  
\end{table*}

\begin{figure*}[t]  
    \centering  
    \includegraphics[width=0.9\linewidth]{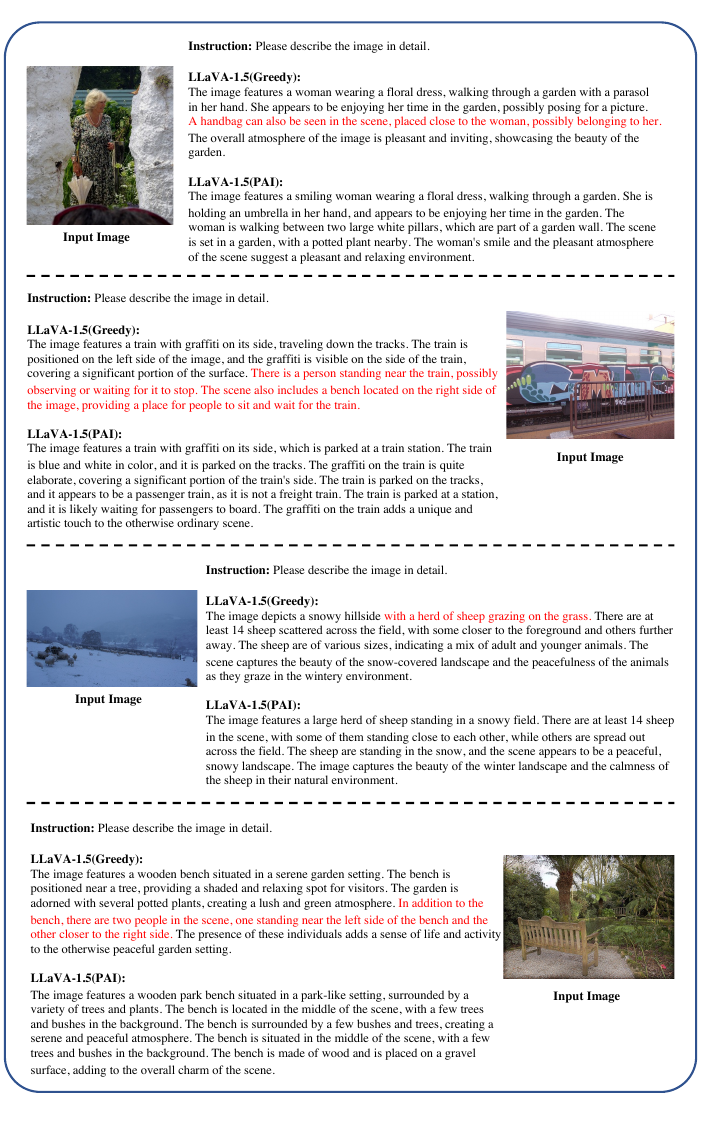}  
    \caption{PAI's performance on reducing hallucinations of LLaVA-1.5-7B.} 
    \label{fig:llava_pai_cases}  
\end{figure*}

\begin{figure*}[t]  
    \centering  
    \includegraphics[width=0.9\linewidth]{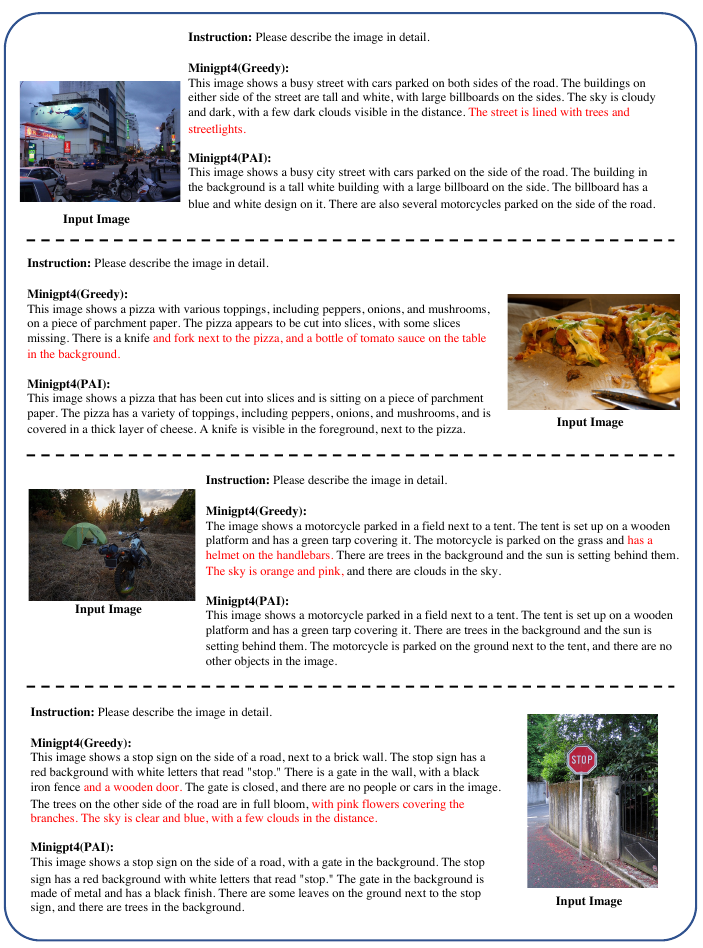}  
    \caption{PAI's performance on reducing hallucinations of Minigpt4.} 
    \label{fig:minigpt4_pai_cases}  
\end{figure*}

\begin{figure*}[t]  
    \centering  
    \includegraphics[width=0.9\linewidth]{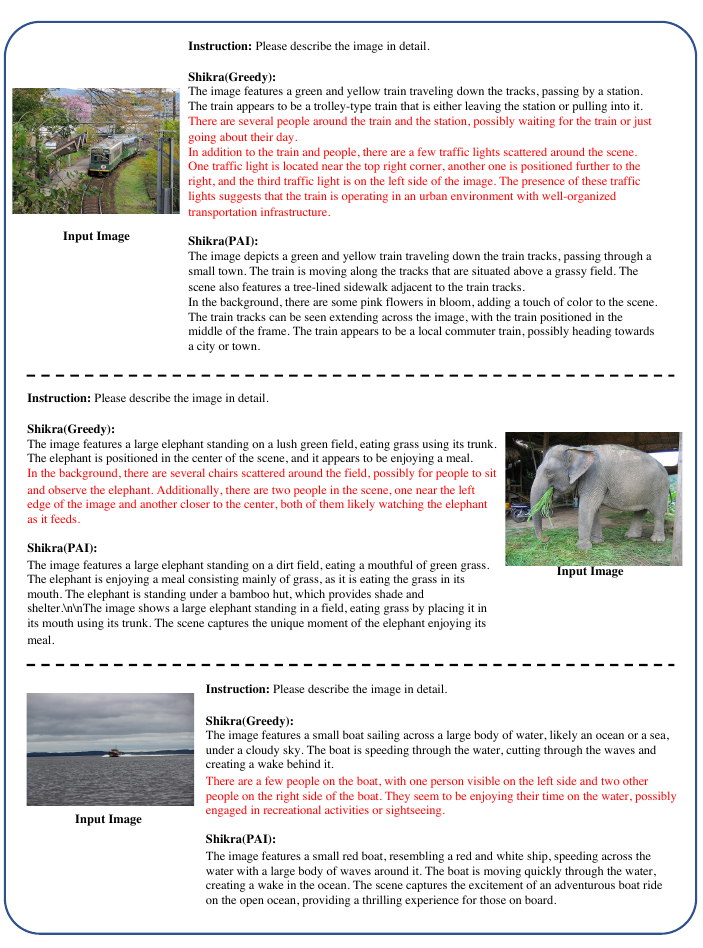}  
    \caption{PAI's performance on reducing hallucinations of Shikra.} 
    \label{fig:shikra_pai_cases}  
\end{figure*}